\documentclass[runningheads]{llncs}
\usepackage[T1]{fontenc}
\usepackage{xcolor}
\usepackage{amsmath,amssymb,amsfonts}
\usepackage{hyperref,booktabs,multirow,array,color,colortbl, float, cleveref }
\usepackage{graphicx,verbatim}
\usepackage{marvosym}
\begin{document}
\bibliographystyle{unsrt}

\title{HGP-Mamba: Integrating Histology and Generated Protein Features for Mamba-based Multimodal Survival Risk Prediction}
\begin{comment}  %% Removed for anonymized MICCAI submission
\author{First Author\inst{1}\orcidID{0000-1111-2222-3333} \and
Second Author\inst{2,3}\orcidID{1111-2222-3333-4444} \and
Third Author\inst{3}\orcidID{2222--3333-4444-5555}}
%
\authorrunning{F. Author et al.}
\institute{Princeton University, Princeton NJ 08544, USA \and
Springer Heidelberg, Tiergartenstr. 17, 69121 Heidelberg, Germany
\email{lncs@springer.com}\\
\url{http://www.springer.com/gp/computer-science/lncs} \and
ABC Institute, Rupert-Karls-University Heidelberg, Heidelberg, Germany\\
\email{\{abc,lncs\}@uni-heidelberg.de}}

\end{comment}

\author{
Jing Dai\inst{1,2} \and
Chen Wu\inst{1,2} \and
Ming Wu\inst{1,2} \and
Qibin Zhang\inst{1,2} \and
Zexi Wu\inst{2} \and \\
Jingdong Zhang\inst{1,2}$^\text{\Letter}$ \and
Hongming Xu\inst{1,2,3}$^\text{\Letter}$
}
\authorrunning{J. Dai et al.}
\institute{Cancer Hospital of Dalian University of Technology, Shenyang, China \\
\email{\{jdzhang\}@cancerhosp-ln-cmu.com} \and
School of Biomedical Engineering, Faculty of Medicine, \\
Dalian University of Technology, Dalian, China \and
Key Laboratory of Integrated Circuit and Biomedical Electronic System, \\
Dalian University of Technology, Dalian, China \\
\email{\{xmu\}@dlut.edu.cn}}

\maketitle
\begingroup                      
  \renewcommand\thefootnote{\Letter}
  \footnotetext{Corresponding author}
\endgroup
\begin{abstract}
Recent advances in multimodal learning have significantly improved cancer survival risk prediction. However, the joint prognostic potential of protein markers and histopathology images remains underexplored, largely due to the high cost and limited availability of protein expression profiling. To address this challenge, we propose HGP-Mamba, a Mamba-based multimodal framework that efficiently integrates histological with generated protein features for survival risk prediction. Specifically, we introduce a protein feature extractor (PFE) that leverages pretrained foundation models to derive high-throughput protein embeddings directly from Whole Slide Images (WSIs), enabling data-efficient incorporation of molecular information. Together with histology embeddings that capture morphological patterns, we further introduce the Local Interaction-aware Mamba (LiAM) for fine-grained feature interaction and the Global Interaction-enhanced Mamba (GiEM) to promote holistic modality fusion at the slide level, thus capture complex cross-modal dependencies. Experiments on four public cancer datasets demonstrate that HGP-Mamba achieves state-of-the-art performance while maintaining superior computational efficiency compared with existing methods. Our source code is publicly available at \url {https://github.com/Daijing-ai/HGP-Mamba.git}.

% Authors must provide keywords and are not allowed to remove this Keyword section.

\end{abstract}
\section{Introduction}
%background
Survival risk prediction is a fundamental task in clinical oncology, aiming to estimate the time to critical events such as death or recurrence and to quantify individual mortality risk. Accurate prediction provides essential insights into disease progression, treatment response, and patient prognosis, ultimately guiding personalized therapeutic decision-making~\cite{wiegrebe2024deep}. 

With the rapid advancement of digital pathology, whole slide images (WSIs), as the gold standard, have been increasingly used in survival risk prediction. WSIs provide a comprehensive view of morphological changes at the cellular and tissue levels, offering a critical basis for survival assessment. Given the large number of patches in WSIs, many methods adopt MIL-based methods for efficient processing and analysis, including attention-based MIL models~\cite{ilse2018attention,yao2020whole,lu2021data} aim to capture global WSI representations. More recently, transformer-based~\cite{jiang2023mhattnsurv,shao2021transmil} and Mamba-based~\cite{gu2024mamba,yang2024mambamil,zhong2025msmmil} architectures leveraging self-attention mechanisms and State Space Models~\cite{kalman1960new} have been introduced, both of which aim to explore mutual-correlations between instances and model long sequence. However, using pathology images alone cannot comprehensively reflect the entire process of cancer occurrence and development. Therefore, the development of robust multimodal approaches is essential yet challenging for constructing accurate and generalizable survival analysis models~\cite{boehm2022harnessing}. With advances in molecular pathology, joint modeling of multimodal data of histology and molecular markers has markedly improved the efficiency and accuracy of survival analysis~\cite{chen2020pathomic,zhou2023cross}. Previous studies on multi-modal fusion largely focused on pairing histopathology images with genomic data, using techniques like cross-attention~\cite{chen2021multimodal,jaume2024modeling} or optimal transport~\cite{xu2023multimodal}.

Although much progress has been achieved, protein markers, which play a critical role in survival risk prediction, have not been fully explored. This is mainly due to the time-consuming and costly assessment of protein markers, which limits their integration into routine clinical workflows~\cite{black2021codex}. Protein markers serve as direct functional mediators of cellular processes and provide unique insights into the tumor microenvironment and molecular signaling pathways, thereby complementing the morphological perspective offered by histology images. Recent studies~\cite{andani2025histopathology,li2026ai} has shown the potential of integrating protein marker prediction into histopathology-based prognostic modeling. Nevertheless, they are limited to representative markers, restricting their ability to provide comprehensive biological insights. Moreover, existing frameworks typically focus on specific cancer types, and their broader utility across other cancer types remains to be verified. Furthermore, large-scale WSIs and high-dimensional proteomics profiles pose significant challenges for efficient fine-grained interaction. The quadratic computational complexity of cross-attention mechanisms often lead to the loss of critical patch-level information and increase the risk of overfitting to task-irrelevant features~\cite{zhang2025me}, which limits its performance and efficiency, making it difficult in clinical settings.

\begin{figure*}[tb]
    \centering
    \includegraphics[width=\columnwidth]{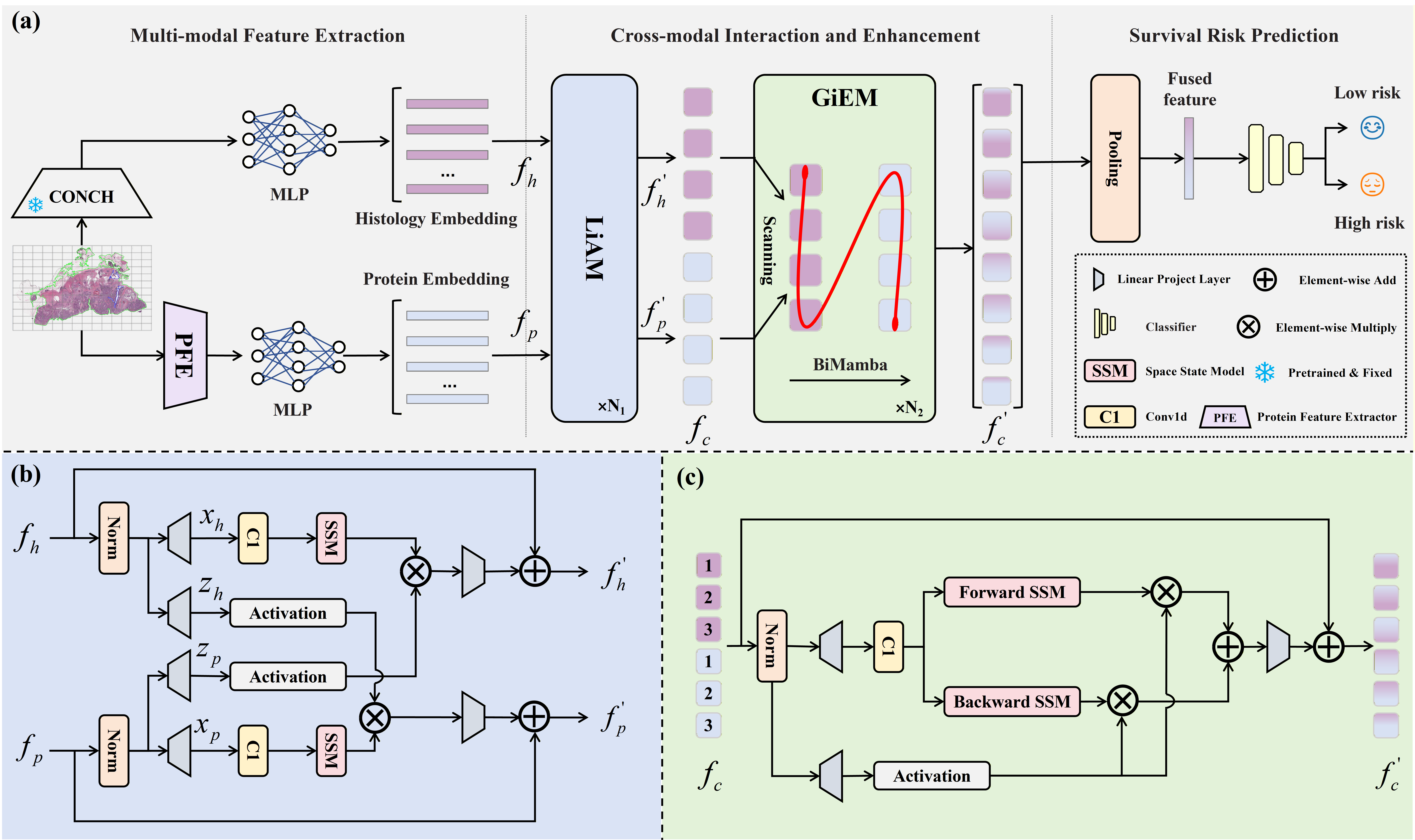}
    \caption{HGP-Mamba overview. (a) Details of the proposed HGP-Mamba which is consisted of three steps: multi-modal feature extraction, feature interaction and enhancement and risk prediction. (b) Schematic of the Local Interaction-aware Mamba (LiAM). (c) Architecture of the Global Interaction-enhanced Mamba (GiEM).}
    \label{fig:overview}
\end{figure*}

To address the above challenges, we propose HGP-Mamba, a Mamba-based framework that captures histology and generated protein features while enabling efficient integration of both modalities. Our model directly derives embeddings for up to 50 protein biomarkers simultaneously from WSIs, enabling reliable multimodal survival modeling even in the absence of measured protein profiles. Furthermore, we introduce a dual-stage Mamba-based fusion mechanism that achieves local cross-modal interaction and global modality cohesion without the substantial computational overhead typically associated with cross attention architectures. HGP-Mamba bridges the gap between tissue morphology and molecular characteristics, providing a cost-effective and scalable alternative to conventional multimodal prognostic frameworks. Our main contributions are summarized as follows:
\begin{itemize}
\item To address the challenges of protein representation learning, we develop a protein feature extractor (PFE) that derives high-throughput protein features directly from WSIs using pretrained foundation models, thereby mitigating data scarcity caused by costly clinical assays.
\item We introduce a hierarchical fusion strategy that comprises a Local Interaction-aware Mamba (LiAM) to capture fine-grained inter-modal dependencies and a Global Interaction-enhanced Mamba (GiEM) to ensure holistic modality integration across the entire tissue landscape.
\item Extensive experiments across four cancer data cohorts demonstrate the superiority and efficiency of the proposed method compared to state-of-the-art baselines in survival risk prediction.
\end{itemize}

\section{Method}
The overall architecture of the proposed HGP-Mamba is shown in Fig.~\ref{fig:overview}, which consists of three main stages: multi-modal feature extraction, cross-modal interaction and enhancement, and survival risk prediction. As depicted in Fig.~\ref{fig:overview}(a), each WSI is divided into thousands of non-overlapping patches. Histology and protein embeddings are then independently extracted using two pretrained foundation models. Subsequently, a Local Interaction-aware Mamba (LiAM) along with a Global Interaction-enhanced Mamba (GiEM) are employed to effectively integrate and enhance histology and protein representations. Finally, the fused multimodal features are used to perform survival risk prediction. In the following sections, we first introduce the preliminary knowledge of state space models (SSMs), and then describe the core components of the proposed HGP-Mamba framework.

\subsection{Space State Models (SSMs)}
SSMs can be viewed as linear time-invariant systems that represent a class of sequence models mapping a one-dimensional input signal \( x(t) \in \mathbb{R} \) to an output response \( y(t) \in \mathbb{R} \) via a latent state \( h(t) \in \mathbb{R}^N \). The system dynamics are formally expressed as:
\begin{equation}
h'(t) = \mathbf{A}h(t) + \mathbf{B}x(t), \quad y(t) = \mathbf{C}h(t),
\label{eq1}
\end{equation}
where $\mathbf{A}\in\mathbb{R}^{N\times N}$ denotes the state transition matrix, and $\mathbf{B}\in\mathbb{R}^{N\times 1}$ and $\mathbf{C}\in\mathbb{R}^{N\times 1}$ denote projection matrices. \( h'(t) \in \mathbb{R}^N \) is the derivative of the hidden state.

For practical implementation, SSMs introduce a discretization timescale parameter $\Delta$ to transform the continuous-time parameters $\mathbf{A}$, $\mathbf{B}$ into their discrete-time counterparts $\mathbf{\bar{A}}$, $\mathbf{\bar{B}}$. The discretization is computed using the Zero-Order Hold (ZOH) method within a specialized CUDA kernel:
\begin{equation}
\mathbf{\bar{A}} = \exp(\Delta \mathbf{A}), \quad \mathbf{\bar{B}} = (\Delta \mathbf{A})^{-1} (\exp(\Delta \mathbf{A}) - I) \cdot \Delta \mathbf{B}
\label{eq2}
\end{equation}

After discretization, the SSM can be written in the discrete-time form: 
\begin{equation}
h_t = \mathbf{\bar{A}} h_{t-1} + \mathbf{\bar{B}} x_t, \quad y_t = \mathbf{C} h_t
\label{eq3}
\end{equation}

Alternatively, the SSM output can be computed using a convolutional formulation, enabling efficient and parallelizable training:
\begin{equation}
\mathbf{K} = (\mathbf{C}\mathbf{\bar{B}}, \mathbf{C}\mathbf{\bar{A}}\mathbf{\bar{B}}, \ldots, \mathbf{C}\mathbf{\bar{A}}^{L-1}\mathbf{\bar{B}}), \quad y = \mathbf{x} \ast \mathbf{K},
\label{eq4}
\end{equation}
where $\ast$ denotes the convolution operation, \( \mathbf{K} \in \mathbb{R}^L \) is an SSM kernel, and \( L \) represents the length of the input sequence $\mathbf{x} = [x(0), \cdots, x(L)]$.

Mamba~\cite{gu2024mamba} further extends SSMs by incorporating selection mechanisms, allowing the model parameters to dynamically depend on the input while leveraging an efficient, hardware-aware parallel algorithm. Consequently, Mamba enables effective and efficient long-sequence modeling by selectively propagating or discarding information along the sequence based on the current token~\cite{yang2024mambamil}.

\subsection{Multi-modal Feature Extraction}
\subsubsection{Histology features} WSIs provide rich morphological information about the tumor microenvironment. However, due to their extremely large size, WSIs cannot be directly processed by convolutional neural networks and need first be partitioned. We begin by segmenting tissue regions, then cut them into $256 \times 256$ non-overlapping patches at $20 \times$ magnification. We employ a pretrained CONCH model~\cite{lu2024visual}, which was initially trained on large-scale WSI-Text pairs, to extract a 512-dimensional embedding for each patch. All patch embeddings from the same WSI are collected as an embedding set. To reduce feature redundancy and computational overhead, we employ a Multi-Layer Perceptron (MLP) to reduce feature dimension and eventually get the histology embedding $f_h\in\mathbb{R}^{N\times D}$, where $N$ denotes the number of patches and $D=256$ is the feature dimension.
\begin{figure}[tb]
  \centering
   \includegraphics[width=1.0\columnwidth]{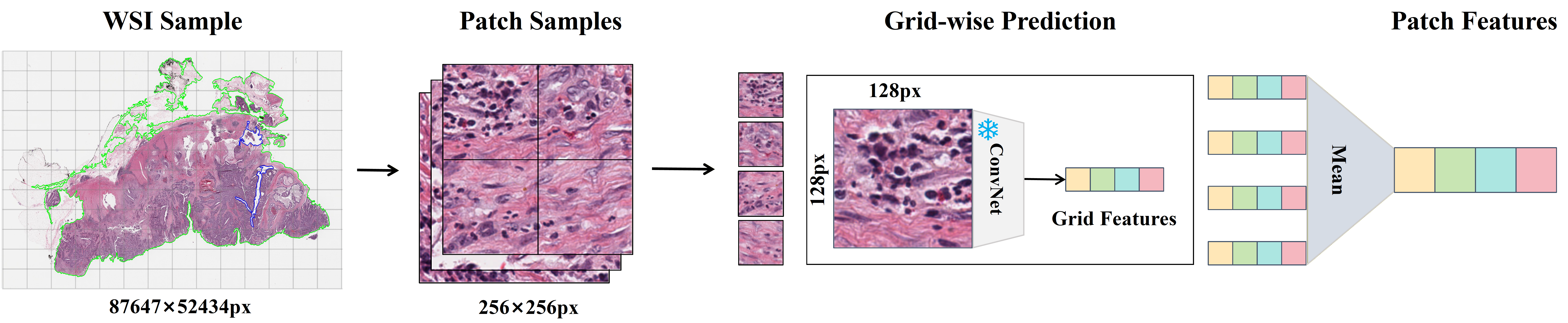}
   \caption{Illustration of our protein feature extractor (PFE). Note that ConvNet is the backbone of the ROISE model.}
   \label{fig:PFE}
\end{figure}
\subsubsection{Protein features} Protein profiles offer valuable insights into molecular biomarkers associated with cancer prognosis. Given the high cost and limited availability of multiplex protein assays, we leverage ROISE~\cite{wu2025rosie}, a pretrained foundation model trained on the largest dataset of co-stained H\&E and immunostained slides. Unlike prior protein prediction frameworks~\cite{andani2025histopathology,li2026ai}, ROISE can simultaneously infer co-expression patterns across up to 50 distinct proteins. Building upon this capability, we develop a Protein Feature Extractor (PFE) to directly derive protein features from WSIs, as illustrated in Fig.~\ref{fig:PFE}. Specifically, each WSI patch is partitioned into non-overlapping grids of size $128 \times 128$. A ConvNet~\cite{liu2022convnet} backbone is then applied to predict expression levels for 50 proteins within each grid. For each marker channel, we compute the mean predicted intensity across all grids, resulting in a normalized  $1 \times M$ feature vector that summarizes the average expression profile at the patch level~\cite{shaban2025foundation}. Finally, a Multi-Layer Perceptron (MLP) projects this representation to a 256-dimensional protein embedding $f_p\in\mathbb{R}^{N\times D}$.

\subsection{Local Interaction-aware Mamba (LiAM)} 
%The high-dimensional nature of WSIs makes it difficult for multi-modal fusion. 
To enable fine-grained cross-modal interaction while maintaining computational efficiency, we introduce a novel LiAM module, illustrated in Fig.~\ref{fig:overview}(b). Given histology and protein features \( f_h\) and \( f_p\), we first obtain their projected representations \( x_h\) and \( x_p\) by applying layer normalization followed by a linear projection. In parallel, \(f_h\) and \(f_p\) are also projected to \(z_h\) and \(z_p\), respectively. The specific formulas are as follows: 
\begin{equation}
\begin{aligned}
\label{x_h, X_p}
x_{m} &= (Linear_{m}^{x}({Norm_{m}}(f_{{m}}))), \quad m \in \{ {h}, {p}\}\\
\end{aligned}
\end{equation}
%------------------------------
\begin{equation}
\begin{aligned}
\label{z_h, z_p}
z_{m} &= (Linear_{m}^{z}({Norm_{m}}(f_{{m}}))), \quad m \in \{ {h}, {p}\} \\
\end{aligned}
\end{equation}
Next, intermediate features \(y_{h}\) and \(y_{p}\) are computed via a 1D convolution, followed by the SiLU activation and a state space model (SSM):
\begin{equation}
\begin{aligned}
\label{y_h, y_p}
y_{m} &= \mathrm{SSM}({SiLU}({Conv_{1d}}({x_{m}}))), \quad m \in \{ {h}, {p}\}\\
\end{aligned}
\end{equation}
To refine cross-modal representations, \(y_{h}\) and \(y_{p}\) are gated by $z_{p}$ and $z_{h}$, respectively, yielding updated outputs \(f_{h}^{\prime}\) and \(f_{p}^{\prime}\):
\begin{align}
\label{output}
f_{h}^{\prime} = y_{h} \odot \mathrm{SiLU}(z_{p}), \quad
f_{p}^{\prime} = y_{p} \odot \mathrm{SiLU}(z_{h})
\end{align}
Finally, residual connections are incorporated into each modality to facilitate gradient flow, thereby improving stability and convergence. Through this interactive learning mechanism, LiAM effectively captures complementary information and models local cross-modal dependencies. 
% More details of the LiAM are elucidated in \textbf{Supplementary Materials}.

\subsection{Golobal Interaction-enhanced Mamba (GiEM)}
Following local cross-modal interaction, we introduce the GiEM module, which employs a bidirectional Mamba (BiMamba)~\cite{zhu2024vision} backbone to further strengthen multimodal representations, as illustrated in Fig.~\ref{fig:overview}(c). Unlike Transformer-based approaches that rely on self-attention to process all tokens simultaneously, GiEM adopts an ordered scanning strategy that preserves the sequential nature of Mamba while enabling efficient global interaction modeling. Specifically, given the multimodal feature sequences, we construct a unified representation $f_c\in\mathbb{R}^{2N\times D}$ by sequentially scanning histology features followed by protein features. This ordered arrangement ensures that information from both modalities is processed in a structured manner, allowing Mamba’s selective scanning mechanism to effectively capture both intra- and inter-modal dependencies.

\subsection{Survival Risk Prediction}
The GiEM module outputs a processed feature sequence $f'_c\in\mathbb{R}^{2N\times D}$, which is then aggregated via max pooling to yield a global feature vector. This vector is subsequently passed into a linear classifier to generate the final survival risk prediction. Following prior studies~\cite{chen2021multimodal,jaume2024modeling}, we simplify the original event time regression problem to a classification problem by dividing the continuous timeline into $n$ intervals. The interval $t_k$ in which the event occurs is used as the class label $k$. The model predicts a hazard vector $H=\left\{h_1, \ldots, h_k, \ldots, h_n\right\}$, where $h_k$ denotes the conditional probability of the event occurring in the $k$-th interval. Each sample is represented as $\{H, c, k\}$, where $c \in\{0,1\}$ indicates the censoring status. The discrete survival function is defined as $f_{\text {surv }}(H, k)=\prod_{i=1}^k\left(1-h_i\right)$. The survival risk prediction loss is formulated as:
\begin{equation}
    \begin{split}
    L_{surv}=&-c\log(f_{sur}(H,k))\\
    &-(1-c)\log(f_{sur}(H,{k-1}))\\
    &-(1-c)\log(h_k).
    \end{split}
\end{equation}

\begin{table*}[tp]
\caption{C-index($\uparrow$) performance(mean ± std) over four datasets. The best and second best results are highlighted in \textcolor{red}{red} and \textcolor{blue}{blue}, respectively.}
\centering
\begin{tabular}{llclclclclc}      
\hline
\multirow{2}{*}{Model} & & \multicolumn{9}{c}{Datasets} \\ \cline{3-11}
 & & COADREAD & & KIRC & & KIRP & & LIHC & & Overall \\ 
\hline
Mean-Pooling & & 0.673$\pm$0.055 & & 0.729$\pm$0.048 & & 0.818$\pm$0.065 & & 0.708$\pm$0.059 & & 0.732 \\
Max-Pooling   & & 0.603$\pm$0.056 & & 0.715$\pm$0.053 & & 0.782$\pm$0.069 & & 0.657$\pm$0.094 & & 0.689 \\
ABMIL~\cite{ilse2018attention}   & & 0.689$\pm$0.071 & & 0.741$\pm$0.045 & & 0.791$\pm$0.099 & & 0.721$\pm$0.075 & & 0.736 \\
CLAM-SB~\cite{lu2021data}                        & & 0.681$\pm$0.041 & & 0.736$\pm$0.055 & & \textcolor{blue} {0.838$\pm$0.046} & & 0.714$\pm$0.083 & & \textcolor{blue} {0.742} \\
CLAM-MB~\cite{lu2021data}                        & & 0.644$\pm$0.069 & & 0.744$\pm$0.036 & & 0.812$\pm$0.075 & & 0.702$\pm$0.070 & & 0.726 \\
TransMIL~\cite{shao2021transmil}         & & \textcolor{red} {0.695$\pm$0.034} & & 0.714$\pm$0.051 & & 0.797$\pm$0.063 & & 0.717$\pm$0.078 & & 0.731 \\
\hline
MambaMIL~\cite{yang2024mambamil}         & & 0.674$\pm$0.082 & & 0.728$\pm$0.065 & & 0.815$\pm$0.034 & & \textcolor{blue} {0.732$\pm$0.060} & & 0.737 \\
BiMambaMIL~\cite{yang2024mambamil}& & 0.683$\pm$0.087 & & 0.739$\pm$0.039 & & 0.796$\pm$0.045 & & 0.729$\pm$0.071 & & 0.737 \\
SRMambaMIL~\cite{yang2024mambamil}& & 0.676$\pm$0.085 & & \textcolor{blue} {0.751$\pm$0.029} & & 0.793$\pm$0.040 & & 0.732$\pm$0.068 & & 0.738 \\
\hline
\rowcolor[rgb]{0.992,0.957,0.945}
HGP-Mamba (Ours)                                      & & \textcolor{blue}{0.695$\pm$0.069} & & \textcolor{red}{0.755$\pm$0.035} & & \textcolor{red}{0.842$\pm$0.054} & & \textcolor{red}{0.739$\pm$0.079} & & \textcolor{red}{0.758} \\
\hline
\end{tabular}
\label{table:comparison}
\end{table*}
\section{Experiments}
\subsection{Datasets and Metrics}
We conducted experiments on four cancer cohorts derived from The Cancer Genome Atlas (TCGA)\footnote{https://portal.gdc.cancer.gov/}, including Colon and Rectum Adenocarcinoma (COADREAD, $n$ = 551), Kidney Clear Cell Carcinoma (KIRC, $n$ = 498), Kidney Papillary Cell Carcinoma (KIRP, $n$ = 261), Liver Cancer (LIHC, $n$ = 311), and Lung Adenocarcinoma (LUAD, $n$ = 455). For cases with multiple slides, one was randomly selected for analysis. We trained the models to predict overall survival (OS) risk and assessed performance using the cross-validated concordance index (C-index)~\cite{heagerty2005survival}, which evaluates how well a survival model ranks patients with their survival time compared to actual survival outcomes.

\subsection{Implementation Details}
We employed 5-fold cross-validation to evaluate our model and other compared methods. We set the number of LiAM blocks $N_1=2$ and the number of GiEM blocks $N_2=1$. Training was conducted for up to 100 epochs with early stopping based on validation C-index. Each epoch used a batch size of 1, with gradients accumulated over 32 steps before backpropagation. The Adam optimizer was used with a learning rate of 2e-4 and a weight decay of 1e-5. During training, a weighted sampling strategy was employed to mitigate class imbalance across all tasks. All experiments were conducted using PyTorch running on a single NVIDIA RTX 4090 GPU.
\begin{figure}[tp]
    \centering
    \includegraphics[width=\columnwidth]{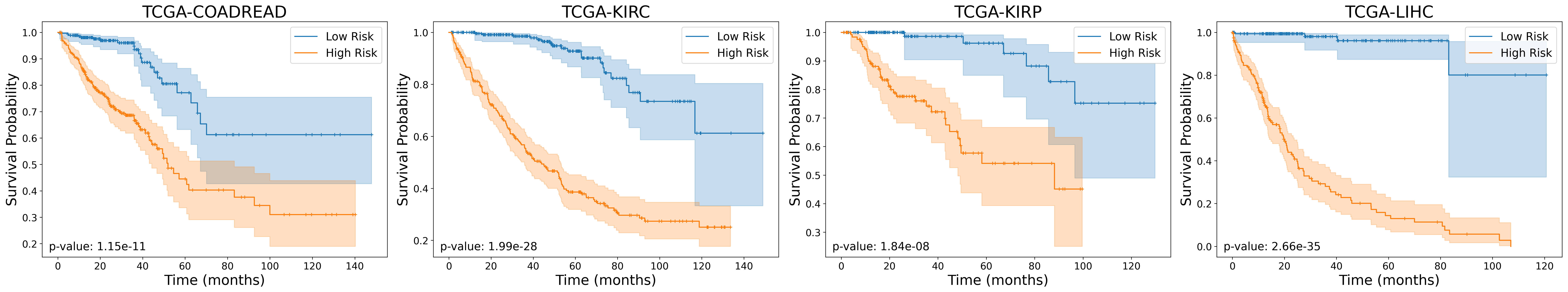}
    \caption{Kaplan-Meier survival curves of the proposed model on four cancer datasets.}
    \label{fig:KM}
\end{figure}
\subsection{Comparison Results}
To demonstrate the effectiveness of HGP-Mamba, we compare it with following methods: \textbf{(1)} conventional pooling methods, including Mean Pooling and Max Pooling; \textbf{(2)} attention-based methods, including ABMIL~\cite{ilse2018attention}, CLAM~\cite{lu2021data} with its two variants CLAM-SB and CLAM-MB; \textbf{(3)} Transformer-based TransMIL~\cite{shao2021transmil}; \textbf{(4)} Mamba-based methods~\cite{yang2024mambamil}, including MambaMIL, BiMambaMIL, and SRMambaMIL. Table~\ref{table:comparison} presents the comparative results, where our proposed HGP-Mamba achieves an average C-index of 75.8\%, surpassing all other comparative methods. Specifically, HGP-Mamba outperforms all prior approaches on the KIRC, KIRP and LIHC datasets and ties for first place with TransMIL on the COADREAD dataset. These results underscore the efficacy of HGP-Mamba in effectively integrating multimodal features and highlight the advantages of multimodal learning for survival prediction.

\subsection{Patient Stratification}
To further validate the effectiveness of HGP-Mamba for survival analysis, we divided all patients into low-risk and high-risk groups according to the median predicted risk scores generated by HGP-Mamba. Kaplan-Meier (KM) analysis was then performed to visualize the survival outcomes across the two groups, as illustrated in Fig.~\ref{fig:KM}. Statistical significance between the risk groups was assessed using the Log-rank test, with a p-value less than 0.05 considered significant. As shown in Fig.~\ref{fig:KM}, all datasets yielded p-values well below 0.05, demonstrating the strong discriminatory ability of HGP-Mamba in survival risk prediction. 

\subsection{Ablation Studies}
\subsubsection{Effectiveness of proposed components} We performed ablation studies in Table~\ref{table:ablation} to evaluate the contributions of the proposed components. First, removing the PFE module reduces HGP-Mamba into a MIL method similar to BiMambaMIL, leading to a significant drop in C-index (e.g., from 0.842 to 0.786 on KIRP). This highlights that the PFE provides prognostic information complementary to raw histological representations.

Next, excluding the LiAM module significantly diminishes the model's capacity to model fine-grained inter-modal interactions, resulting in a notable decline in C-index. LiAM utilizes a gating mechanism that dynamically weighs the interaction between histology and protein features, which can learn to suppress noisy influence during the training process. This design ensures robust survival risk prediction. 

Finally, omitting the GiEM module degraded modality cohesion and produced suboptimal performance. Overall, these ablations confirm that PFE, LiAM, and GiEM each play essential and complementary roles. Together, they provide a comprehensive representation of tumor morphology and molecular heterogeneity, which is essential for accurate survival risk prediction.
\begin{table}[tp]
\caption{Ablation study assessing the impact of three key modules in HGP-Mamba.}
\resizebox{\columnwidth}{!}{
\centering
\setlength{\extrarowheight}{0pt}
\addtolength{\extrarowheight}{\aboverulesep}
\addtolength{\extrarowheight}{\belowrulesep}
\setlength{\aboverulesep}{0pt}
\setlength{\belowrulesep}{0pt}
\begin{tabular}{lcccc} 
\hline
\multirow{2}{*}{Configuration} & \multicolumn{4}{c}{Datasets} \\ \cline{2-5}
                               & COADREAD & KIRC & KIRP & LIHC \\ 
\hline
w/o PFE & 0.687 $\pm$ 0.064 & 0.739 $\pm$ 0.027 & 0.786 $\pm$ 0.119 & 0.715 $\pm$ 0.087 \\
w/o LiAM & 0.629 $\pm$ 0.067 & 0.751 $\pm$ 0.045 & 0.781 $\pm$ 0.109 & 0.711 $\pm$ 0.083 \\
w/o GiEM & 0.681 $\pm$ 0.071 & 0.744 $\pm$ 0.045 & 0.800 $\pm$ 0.082 & 0.727 $\pm$ 0.079 \\
\rowcolor[rgb]{0.992,0.957,0.945}
Full (HGP-Mamba) & \textbf{0.695 $\pm$ 0.069} & \textbf{0.755 $\pm$ 0.035} & \textbf{0.842 $\pm$ 0.054} & \textbf{0.739 $\pm$ 0.079} \\
\hline
\end{tabular}
}
\label{table:ablation}
\end{table}

\begin{figure}[tp]
    \centering
    \includegraphics[width=\columnwidth]{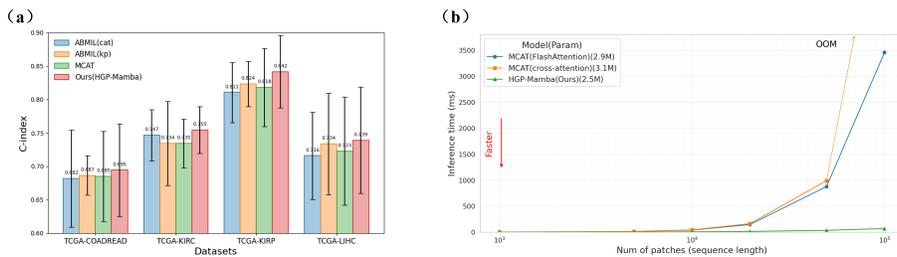}
    \caption{(a) Comparison of different multimodal fusion methods. (b) Inference time comparison with varying patches}
    \label{fig:ab2}
\end{figure}

\subsubsection{Superiority of multimodal fusion method} To further demonstrate the superiority of our multimodal interaction and enhancement modules, we replaced LiAM and GiEM with several baseline fusion method for survival risk prediction. First, we incorporated two common late-fusion schemes using ABMIL with feature concatenation (ABMIL-Cat)~\cite{mobadersany2018predicting}, and ABMIL with a Kronecker Product fusion (ABMIL-KP)~\cite{chen2022pan}. As shown in Fig.~\ref{fig:ab2}(a), both approaches underperform relative to HGP-Mamba, indicating their limited ability to capture complex tumor–microenvironment relationships. We also compared HGP-Mamba with the leading early fusion method, MCAT~\cite{chen2021multimodal,andani2025histopathology,li2026ai}. Because of GPU memory constraints, we substituted MCAT’s cross-attention module with FlashAttention~\cite{dao2022flashattention} in our implementation. Our method leverages the dual-stage Mamba architecture, achieving superior predictive performance while preserving high computational efficiency.

Morever, we conducted efficiency analysis using parameter count and inference time. For fairness, the comparison focuses exclusively on the cross-modal interaction and enhancement stage and excludes the cost of multimodal feature extraction; all experiments were executed under identical conditions. We benchmarked HGP-Mamba against the Transformer-based MCAT implemented with its two core attention mechanisms (co-attention and FlashAttention). Specifically, we constructed histology embedding sequences of length 1,000, 5,000, 10,000, 20,000, 50,000, and 100,000 with an embedding dimension of 512, while keeping protein embeddings fixed at dimension 50.

As shown in Fig.~\ref{fig:ab2}(b), HGP-Mamba has a substantially smaller parameter footprint (2.47 MB) than the Transformer baselines and consistently achieves large speedups across all sequence lengths. For example, when processing 50,000 tokens, HGP-Mamba requires only 34.08 ms, representing a 96.1\% and 96.6\% reduction in inference time compared with FlashAttention-based MCAT (875.26 ms) and co-attention MCAT (988.73 ms), respectively. These results indicate that HGP-Mamba not only maintains stable predictive performance for survival risk but also offers a major advantage in computational efficiency for cross-modal interaction.

\begin{figure}[tb]
  \centering
   \includegraphics[width=1.0\columnwidth]{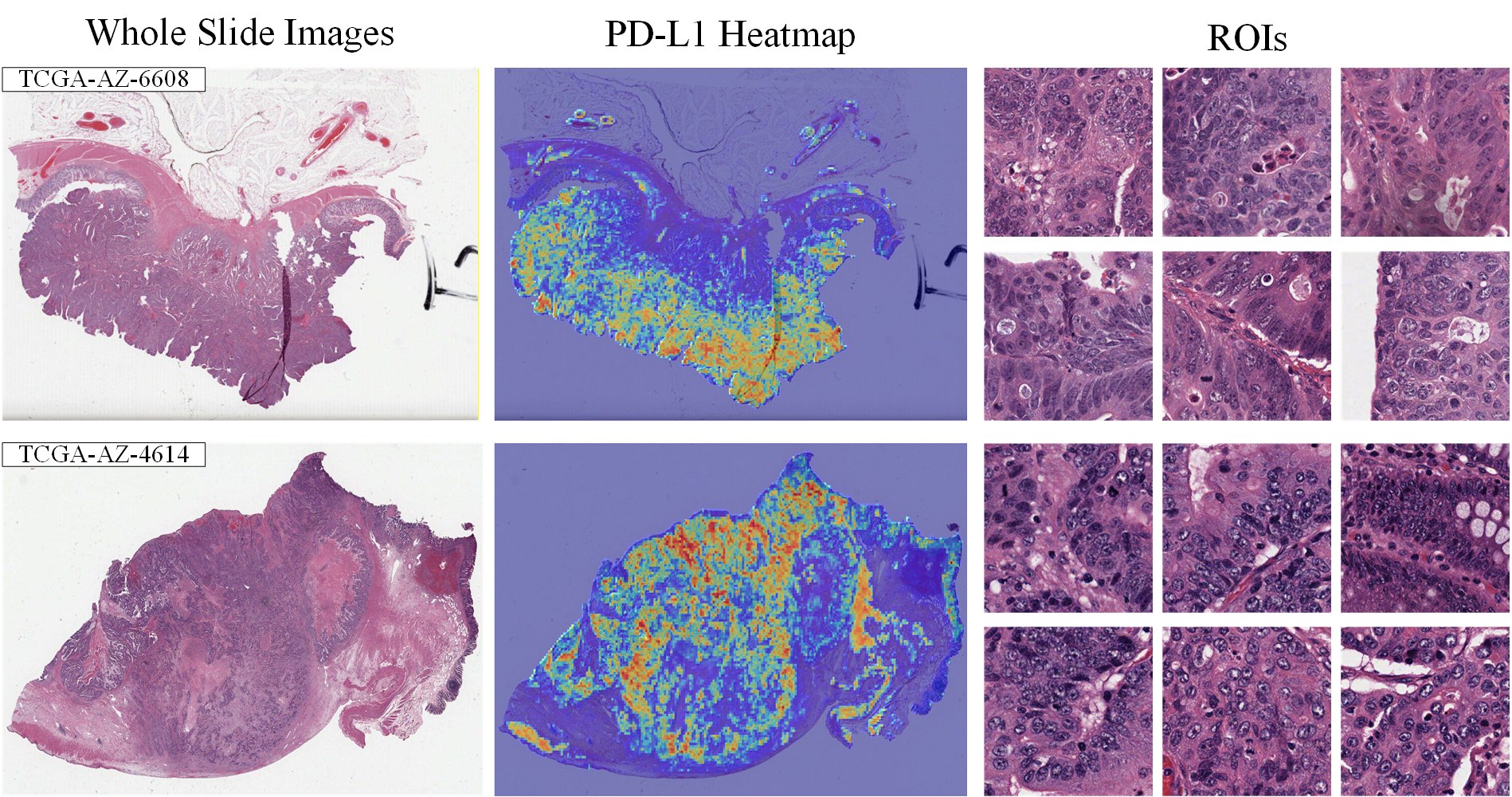}
   \caption{Spatial expression heatmaps for PD-L1 on randomly selected slides from the TCGA-COADREAD dataset. For each sample, the left panel shows the WSI thumbnail, the middle panel overlays the predicted PD-L1 expression heatmap on the WSI, and the right panel displays the selected patches according to the predicted expression level.}
   \label{fig:visualization}
\end{figure}

\subsection{Protein Visualization}
To validate whether the PFE captures authentic molecular signals, we visualized protein expression patterns (e.g., PD-L1) across the TCGA-COADREAD cohort. As shown in Figure~\ref{fig:visualization}, the predicted high-expression areas (red) exhibit strong spatial concordance with histologically confirmed tumor regions. Detailed ROI analysis further demonstrates that the inferred signals are confined to specific biological compartments, such as tumor nests and stroma with immune infiltration. This consistency with pathological ground truth suggests that HGP-Mamba successfully bridges the gap between raw morphology and underlying molecular heterogeneity.

\section{Conclusion}
In this work, we propose HGP-Mamba, a Mamba-based framework that captures histology and generated protein features while enabling efficient integration of both modalities. By leveraging pretrained foundation models, HGP-Mamba directly extracts high-throughput protein features from WSIs, thereby mitigating the scarcity of measured protein profiles. Through Mamba-based cross-modal interaction and enhancement, the framework effectively captures tumor heterogeneity and yields a more comprehensive representation for cancer survival risk prediction. Given its superiority and efficiency, HGP-Mamba can be extended to more complex tasks involving diverse data modalities, facilitating future translation toward clinical applications.

\bibstyle{splncs04}
\bibliography{ref}

\end{document}